\documentclass[letterpaper, 10 pt, conference]{ieeeconf} 
\IEEEoverridecommandlockouts
\usepackage{cite}
\usepackage{amsmath,amssymb,amsfonts}
\usepackage{algorithmic}
\usepackage{graphicx}
\usepackage{textcomp}
\usepackage[table, dvipsnames]{xcolor}
\usepackage{caption}
\usepackage{float}
\usepackage{subcaption}
\usepackage{tabularray}
\usepackage{multirow}
\usepackage{booktabs}
\usepackage{hyperref}
\usepackage{titlesec}
\usepackage[disable]{todonotes}
\DeclareCaptionFont{mysize}{\fontsize{8}{10}\selectfont}
\usepackage{soul}
\usepackage[subtle,tracking=normal]{savetrees}

\def\BibTeX{{\rm B\kern-.05em{\sc i\kern-.025em b}\kern-.08em
    T\kern-.1667em\lower.7ex\hbox{E}\kern-.125emX}}

\newcommand{\multilinecomment}[1]{}

\author{Zeynep Özge Orhan$^{1,2}$, Andrea Dal Prete$^{1,3}$, Anastasia Bolotnikova$^{2,4}$,\\ Marta~Gandolla$^{3}$,  Auke Ijspeert$^{2}$, and Mohamed Bouri$^{1}$ 
\thanks{$^{1}$ Zeynep Özge Orhan, Andrea Dal Prete, Mohamed Bouri are with the REHAssist group,  Ecole Polytechnique Federale de Lausanne (EPFL). {\tt\footnotesize (e-mail: zeynep.orhan@epfl.ch, andrea.dalprete@mail.polimi.it, mohamed.bouri@epfl.ch)}  }
\thanks{$^{2}$ Zeynep Özge Orhan, Anastasia Bolotnikova and Auke Ijspeert are with the BioRobotics Laboratory, Ecole Polytechnique Federale de Lausanne (EPFL). {\tt\footnotesize (e-mail: zeynep.orhan@epfl.ch, anastasia.bolotnikova@epfl.ch, auke.ijspeert@epfl.ch)}  }
\thanks{$^{3}$  Andrea Dal Prete and Marta~Gandolla are with Politecnico di Milano, Department of Mechanical Eng. {\tt\footnotesize (e-mail: andrea.dalprete@mail.polimi.it, marta.gandolla@polimi.it)}  }
\thanks{$^{4}$ Anastasia Bolotnikova is with the Reconfigurable Robotics Lab, Ecole Polytechnique Federale de Lausanne (EPFL). {\tt\footnotesize (e-mail: anastasia.bolotnikova@epfl.ch)} }
}

\begin{document}

\title{Maximizing Performance with Minimal Resources\\ for Real-Time Transition Detection}

\maketitle

\begin{abstract}
Assistive devices, such as exoskeletons and prostheses, have revolutionized the field of rehabilitation and mobility assistance. Efficiently detecting transitions between different activities, such as walking, stair ascending and descending, and sitting, is crucial for ensuring adaptive control and enhancing user experience. We here present an approach for real-time transition detection, aimed at optimizing the processing-time performance. By establishing activity-specific threshold values through trained machine learning models, we effectively distinguish motion patterns and we identify transition moments between locomotion modes. This threshold-based method improves real-time embedded processing time performance by up to 11 times compared to machine learning approaches. The efficacy of the developed finite-state machine is validated using data collected from three different measurement systems. Moreover, experiments with healthy participants were conducted on an active pelvis orthosis to validate the robustness and reliability of our approach. The proposed algorithm achieved high accuracy in detecting transitions between activities. These promising results show the robustness and reliability of the method, reinforcing its potential for integration into practical applications.
\end{abstract}


\section{Introduction}

Locomotion mode identification involves recognizing various human activities, such as walking, running, stair ascent, stair descent, and sitting \cite{Kolaghassi2021}. This field has significantly evolved over the years, with a focus on achieving accurate human intention recognition.

Wearable sensors like inertial measurement units (IMUs), force sensors, and electromyography (EMG) sensors have played a pivotal role in locomotion-mode identification \cite{Gehlhar2023}. Researchers have commonly used feature extraction methods to capture activity-specific characteristics, followed by classification algorithms to identify these activities \cite{Allahbakhshi2019, Aldabbagh2020}. Classification methods vary in complexity, ranging from simple threshold-based (TH) approaches to advanced deep learning techniques \cite{Preece2009, Wang2023}.

\begin{figure}[t]
    \centering
    \includegraphics[width=0.75\linewidth]{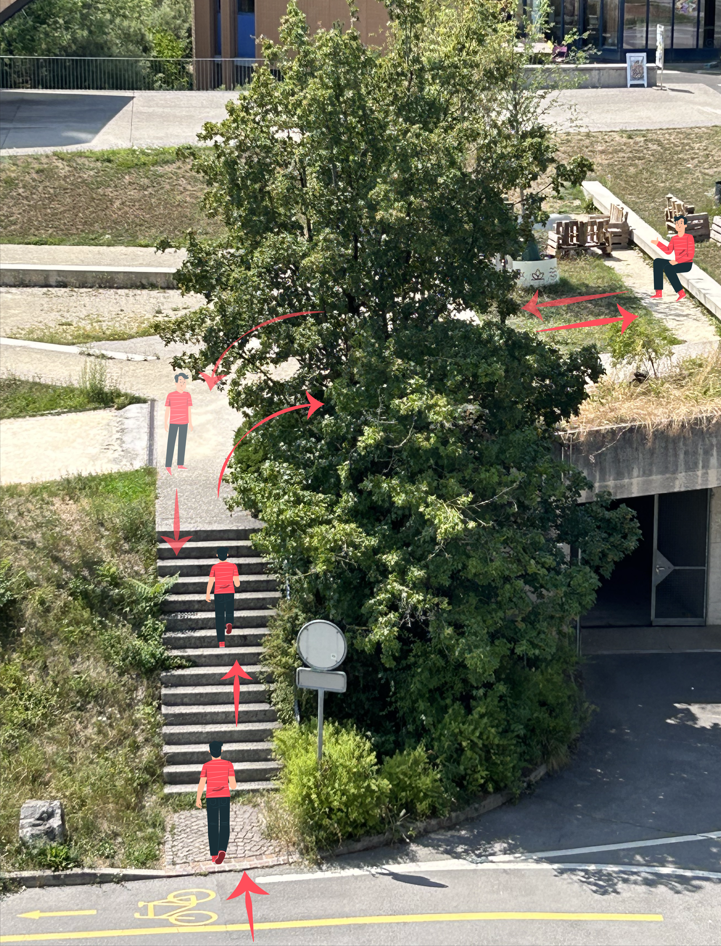}
    \vspace{-0.2cm}
\caption{The outdoor environment where the onboard transition detection tests are performed for walk-stair ascend, stair ascend-walk, walk-sit, sit-walk, walk-stair-descend, and stair descend-walk transitions}
\label{fig:OnlineTestsOutdoorEnvironment}
\end{figure}

Reported accuracy varies across different studies, and depends on factors such as choice of sensors, feature extraction methods, and classification algorithms \cite{Attal2015, Pohl2022}. Most of the time, the performance of the selected techniques is analyzed based on accuracy, sensitivity, and specificity. However, for onboard real-time implementation of suggested strategies, the low computational cost is also important \cite{Yuan2015}. Since the performance of an assistive device is strongly affected by the response time or delay, an accurate but also fast transition detection strategy is essential \cite{LoboPrat2014, Aldabbagh2020, Kang2020}. 

Existing classification methods excel in identifying steady-state activities for various locomotion modes \cite{Donahue2022, Gehlhar2023}. However, detecting the transitions between different activities is crucial from the perspective of control of a lower-limb exoskeleton to be able to assist promptly with changes in locomotion, ensuring smooth human-robot interaction and enhancing the safety and efficiency of lower-limb exoskeletons. Hence, a need to develop specialized algorithms that focus on transition detection rather than continuous activity recognition.

Most of the studies primarily rely on the sensors placed on the human body and often lack real-time experimental results integrated into lower-limb assistive devices \cite{Zhou2019, Vu2022}. There are few studies with exoskeletons that conducted experiments followed by offline recognition and analysis and even less with online recognition \cite{Labarriere2020}. 

In the field of lower limb prosthesis, multiple methodologies are developed to achieve high real-time performance in recognizing different locomotion types for intuitive prosthetic control \cite{Young2014b, Su2019}. However, the identification of transitions between locomotion modes with lower-limb exoskeletons has not been studied as comprehensively in prosthetic systems. In the review article \cite{Labarriere2020}, it is shown that only three studies investigated the transitions between locomotion modes among the eight studies focused on orthoses or exoskeletons. In a more recent review \cite{Moreira2022}, the number of studies is increasing to detect transitions between locomotion modes.

In Long et al., \cite{Long2016}, the Support Vector Machine (SVM) algorithm is optimized by particle swarm optimization to identify five locomotion modes with a knee exoskeleton: level-ground walking, stair ascent/stair ascent, ramp ascent/descent, and transitions between them based on ground reaction forces and attitude and heading reference system. Human subject experiments achieved low error rates, but the suggested system may not translate well to seamless online control since the identification of transitions between certain modes relies on the entire upcoming step.

In Wang et al., an LSTM-based model is implemented on a soft knee exoskeleton to detect the level walking, stair ascent/descent, and the transitions between them based on IMUs \cite{Wang2018}. They report 98.2$\%$ for steady-state accuracy while the recognition delay of all transitions is slightly less than one step. Zhou et al. also studied the same three locomotion modes and transitions between them \cite{Zhou2019} by using SVM with the data of IMUs for a unilateral knee exoskeleton. However, these approaches \cite{Wang2018, Zhou2019} rely on fixed-length sliding time windows that shift over the gait time series, potentially limiting its adaptability to variable walking speeds. 

Liu and Wang, \cite{Liu2020} implemented SVM to recognize the locomotion mode for a unilateral knee exoskeleton together with recognition of sitting. Their experimental results demonstrated that the real-time recognition with two IMUs obtained satisfactory recognition accuracy and low delay time. A fuzzy logic-based algorithm for locomotion and transition mode recognition based on the uses of inertial sensors mounted on a hip joint exoskeleton is proposed in \cite{Du2021}. The proposed approach is able to overcome the subject-dependent parameters in data training, avoiding a training procedure per subject. In contrast, subject-independent locomotion mode classification for a hip exoskeleton is suggested by \cite{Kang2022} based on a deep CNN. Although this suggested strategy seems to be promising without fine-tuning subject-specific parameters, it has only been validated in offline scenarios.

TH methods offer transparency and interpretability, allowing clear rules and decision-making criteria. Since they are computationally efficient, they are suitable for real-time applications. However, they may struggle with complex patterns, have limited generalization, and require manual threshold creation. Most of the traditional machine learning approaches can handle complicated patterns, generalize well, and provide versatility with various algorithms. Yet, they often require feature engineering and are sensitive to feature quality, limiting scalability. On the other hand, deep learning approaches automatically learn features from raw data, excel at capturing complex patterns, and scale well. However, they need large amounts of labeled data, are computationally demanding, and lack interpretability. 

By considering the data availability and limited computational resources on assistive devices, a TH method is proposed in this work. Instead of setting thresholds via manual inspection, traditional machine learning approaches are utilized considering the generalizability of the approach. 

Therefore, this study will focus on recognizing transitions between locomotion types such as overground walking, sitting, slope ascend/descend, and stair ascend/descend, while minimizing the computational resources on the required processing power and time on lower-limb assistive devices for real-time onboard implementation. 

The major contributions of this study are the following.
\begin{itemize}
    \item The proposed threshold-based implementation offers a significant advancement in real-time embedded performance, reducing computational costs while maintaining high accuracy in real-time classification. By utilizing conventional machine learning techniques, the threshold-setting process is simplified, resulting in a more efficient and streamlined approach. 
    \item The efficacy of the developed finite-state machine is validated using data collected from three different measurement systems: i) a Vicon camera motion capture system, ii) goniometers, and force plates, and iii) an IMUs-based motion capture system. This comprehensive validation ensures the reliability and versatility of the proposed approach across different measurement modalities, enhancing its applicability in practical scenarios.
    \item Robustness and reliability of the suggested strategy to use in lower-limb wearable robots are verified through the human subject experiments with an active hip orthosis with a total of 10 healthy participants.
\end{itemize}
\smallskip

\smallskip
\section{Methods}
\label{sec:Methods}

\subsection{Training Dataset}
\label{sec:Method-Dataset}
Human gait kinematics and kinetics are studied for various distinct locomotion scenarios in the literature together with several public datasets \cite{Reiss2012, Zhang2012, Fukuchi2018, Khandelwal2017, Chereshnev2018, Horst2021, Luo2020}. However, in most of the datasets, transitions between activities are not considered. To be able to detect the transitions, the public dataset provided by \cite{Reznick2021} is used to train the classifiers. This dataset involves several activities including walking, stair ascending/descending, sitting, and standing with continuous variations and transitions between activities. The transitions between different locomotion modes are notated as Walk to Sit (W $\rightarrow$ S), Sit to Walk (S $\rightarrow$ W), Walk to Stair Ascend (W $\rightarrow$ SA), Stair Ascend to Walk (SA $\rightarrow$ W), Walk to Stair Descend (W $\rightarrow$ SD), Stair Descend to Walk (SD $\rightarrow$ W) and finally all the steady state cases are notated with SS-\textit{mode}.


Gait data were acquired with a 10-cameras Vicon T40 motion capture system on ten healthy subjects. The training dataset is constructed as described in \cite{Cheng2021}.
Of the available data, 72\% of samples are taken for the training while 18\% of samples are used for tests of each individual classifier. The remaining 10\% of data is later used for validation of the method together with FSM.

\subsection{Features and State-Machine}

In \cite{Cheng2021}, they derive kinematics-based classification features called instantaneous characteristic features (ICFs) based on thigh angle ($\theta$) and thigh velocity ($\dot{\theta}$). Three ICFs are derived for transitions between Walk and Sit, Walk and Stair Ascent, and Walk and Stair Descent at three specific moments. The specific moments are defined as maximum hip flexion (MHF), heel strike (HS), and $\theta_{10}$ (TH10). The MHF and HS are used for locomotion transition classifications while TH10 is used for walk and sit transition. 
ICF-1 is $\theta_{th}$ at MHF $\theta_{th,MHF}$, ICF-2 is the difference $\theta_{th,MHF}$ - $\theta_{th,HS}$ and ICF-3 is the sign of $\dot{\theta}_{th}$, when $\theta_{th} = 10^\circ$ as shown in Tab.~\ref{tab:ICFDefinitions}.

\begin{table}[t]
\centering
\caption{The derived features for transitions between distinct locomotion modes}
\label{tab:ICFDefinitions}
\begin{tabular}{lll}
\toprule
                        & Notation & Definition\\ 
\midrule
W $\rightarrow$ S      & ICF-3 & The sign of, $\dot{\theta}_{th}$, when $\theta_{th} = 10^\circ$\\ 
S $\rightarrow$ W      & ICF-3 & The sign of, $\dot{\theta}_{th}$, when $\theta_{th} = 10^\circ$\\ 
SA $\rightarrow$ W      & ICF-1 & $\theta_{th}$ at MHF $\theta_{th,MHF}$\\ 
W $\rightarrow$ SA     & ICF-1 & $\theta_{th}$ at MHF $\theta_{th,MHF}$\\ 
W $\rightarrow$ SD     & ICF-2 & The difference $\theta_{th,MHF}$ - $\theta_{th,HS}$\\ 
SD $\rightarrow$ W      & ICF-2 & The difference $\theta_{th,MHF}$ - $\theta_{th,HS}$\\ 
\bottomrule
\end{tabular}
\vspace{-0.2cm}
\end{table}

\begin{figure}[b]
    \centering
    \includegraphics[width=0.85\linewidth]{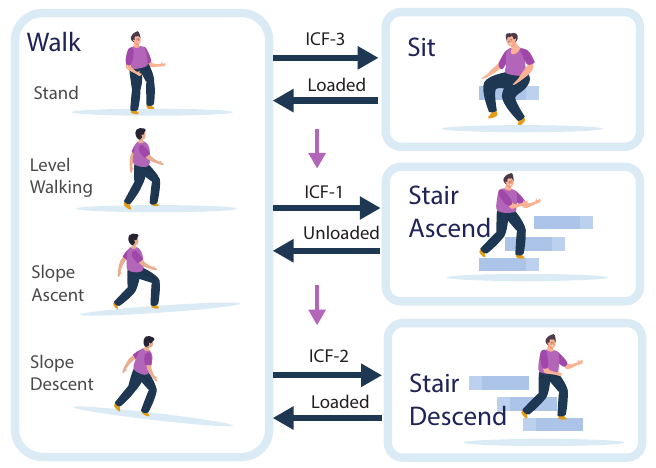}
    \vspace{-0.2cm}
\caption{The finite state machine suggested by Cheng et al. in with the updated value of thigh boundary for ICF3 \cite{Cheng2021}.}
\label{fig:fsm}
\end{figure}

 In \cite{Cheng2021}, the researchers employ a finite state machine (FSM) comprising four distinct states — Sit, Walk, Stair Ascent, and Stair Descent — to decompose the classification challenge into six transition detection problems, each corresponding to the current activity. We are utilizing the same FSM introduced by \cite{Cheng2021}, as depicted in Fig.~\ref{fig:fsm}. Within the FSM, level walking, standing, and ramp walking are grouped into a unified state.  In the flowchart, the ICFs detector will check for identifying distinct moments (e.g., MHF, HS) containing various ICFs from the thigh angle and ground reaction forces to predict the transitioning state.

\subsection{Classification Algorithms}
Several ML algorithms such as Linear Support Vector Machines (SVM), Logistic Regression (LR), k-Nearest Neighbors, Naïve Bayes, Random Forest, and Gradient Boosting are used to train the classifiers. After manually tuning the parameters based on the accuracy results of each algorithm, the best-performing one in terms of training accuracy is selected. Training and test accuracy for the best performing approach for each transition are reported in Table~\ref{tab:TrainingAccuracy}. 

\todo[cross validation??]

\begin{table}[t]
\centering
\caption{Training and test accuracies with selected algorithms for each classifier}
\label{tab:TrainingAccuracy}
\begin{tabular}{llll}
\toprule
                        & Algorithm        & Training Acc.      & Test Acc. \\ 
\midrule
W$\rightarrow$S              & Linear SVM        & 86.94\%           & 90.91\% \\ 
S$\rightarrow$W              & LR                & 95.07\%           & 100.0\% \\ 
W$\rightarrow$SA   & LR                & 91.44\%           & 89.36\% \\ 
SA$\rightarrow$W     & LR                & 81.67\%           & 86.89\% \\ 
W$\rightarrow$SD   & LR                & 93.20\%           & 94.38\% \\ 
SD$\rightarrow$W    & LR                & 97.25\%           & 96.55\% \\ 
\bottomrule
\end{tabular}
\vspace{-0.4cm}
\end{table}

\subsection{Threshold-based Implementation}

Many of the assistive devices are resource-constrained embedded systems to perform real-time operations. For assistive devices, it is critical to be computationally efficient in the context of real-time embedded systems since it enables quick and responsive control actions. To improve computational efficiency and optimize resource usage, we are introducing TH implementation for a practical and efficient solution. The computational efficiency, ease of implementation, interpretability, and inherent robustness of a TH method, make it an appealing choice for transition detection in the control of assistive devices.


In the method proposed in \cite{Cheng2021}, machine learning models were trained using 1-D feature space and resulted in thresholds on ICFs for various transition activities. However, deploying these classifiers on devices can introduce additional computational costs, potentially causing delays in the system, especially in resource-constrained environments. By performing offline training, a threshold value is defined for each classifier. Therefore, the ML models can be substituted by a straightforward $if/else$ condition based on the trained thresholds while maintaining accuracy and ensuring efficient performance in real-time embedded operation. 
The check for the classifiers shown in Fig.~\ref{fig:fsm}, is therefore implemented via thresholds.

\section{Experiments}
\label{sec:Experiments}

\subsection{Offline Experiments}
\subsubsection{Testing on two different data sets}
To evaluate the performance of the trained models, the classifiers are tested with FSM by using two different human gait datasets. To simulate the performance in real-time implementation, gait data is provided to the FSM consecutively "sample-by-sample" with the goal of mimicking upcoming sensor data. Therefore, the detection of ICF, extraction of the corresponding feature, and performing the binary classification are tested simultaneously in this evaluation. 

The first data is the 10\% of the dataset provided by \cite{Reznick2021} that has been described in Section~\ref{sec:Method-Dataset}. The second gait dataset \cite{Camargo2021} contains gait data from 22 healthy subjects for several locomotion modalities such as level-ground and treadmill walking, stair ascent, stair descent, slope ascent, and slope descent. Although multiple locomotion modes at different conditions are available in this dataset, walk-to-sit and sit-to-walk transitions were not included. Therefore the performance of sit-to-stand and stand-to-sit transitions aren't taken into account. 

The accuracy results for the two datasets \cite{Reznick2021, Camargo2021} are presented in Table~\ref{tab:RealTimeAcuracy}. Data from \cite{Reznick2021} is also used for testing and validation of the trained model. However, as stated in Section~\ref{sec:Method-Dataset}, $18\%$ of the data is separated for this experiment.

\begin{table}[h]
\centering
\caption{Transition detection accuracy with two different datasets}
\label{tab:RealTimeAcuracy}
\begin{tabular}{lccl}
\toprule
                        & Data from \cite{Reznick2021}        & Data from \cite{Camargo2021}  \\ 
\midrule
W$\rightarrow$S             & 90.00\%                & - \%  \\ 
S$\rightarrow$W             & 99.90\%                & - \%  \\ 
W$\rightarrow$SA     & 90.00\%                & 99.90\%  \\ 
SA$\rightarrow$W    & 80.33\%                & 93.33\%  \\ 
W$\rightarrow$SD    & 90.00\%                & 80.00\%  \\ 
SD$\rightarrow$W    & 85.00\%                & 80.00\% \\ 
\bottomrule
\end{tabular}
\vspace{-0.2cm}
\end{table}


\subsubsection{Results with IMU-based motion capture system}

\begin{figure}[t]
    \centering
    \includegraphics[width=0.61\linewidth]{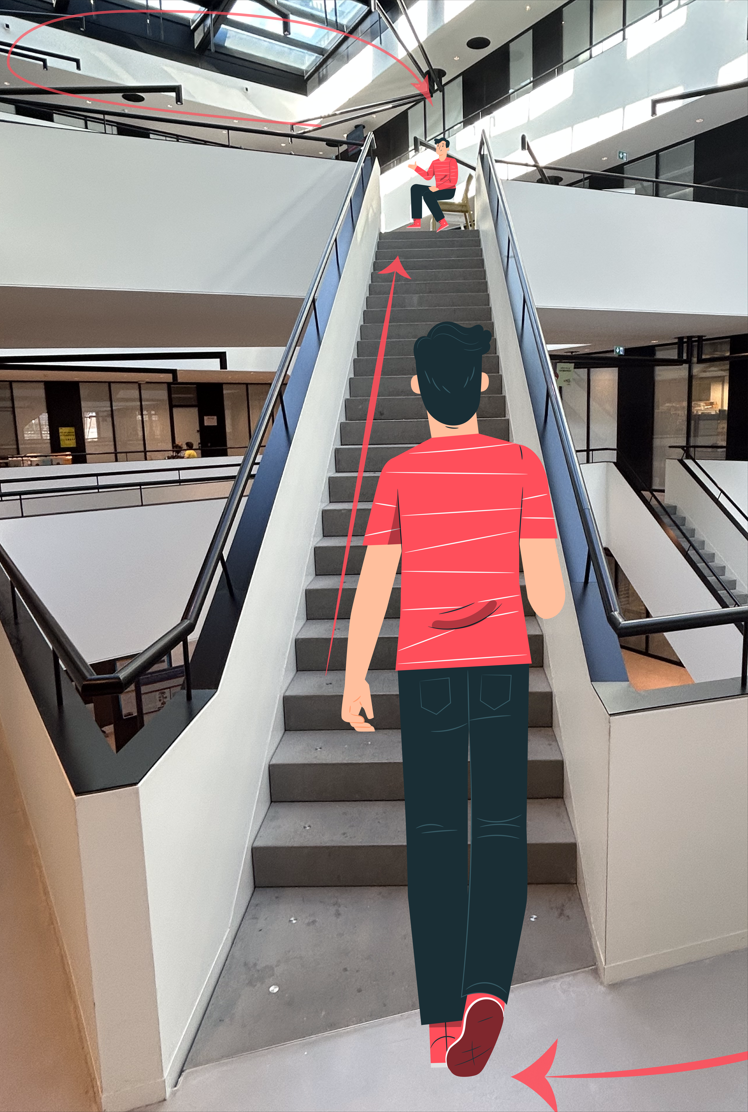}
    \vspace{-0.2cm}
\caption{The indoor environment where the test is performed with an IMU-based motion capture system}
\label{fig:XsensScenario}
\end{figure}

An offline test was conducted in an indoor environment with a single subject to assess the method's suitability for real-time implementation. The subject was a 25-year-old female with a height of 167 cm and a weight of 65 kg. Kinematic data during human gait were captured using the MVN Awinda motion capture system by Xsens (Xsens Technologies, B.V., The Netherlands) \cite{Roetenberg2009}, which utilizes IMUs to accurately track motion.

To evaluate the classifiers, a predefined protocol was followed in a controlled environment, incorporating sitting, overground walking, and stair ascend/descend activities as shown in Fig.~\ref{fig:XsensScenario}. The data collection procedure is followed the following transitions: S$\rightarrow$W, W$\rightarrow$SD, SD$\rightarrow$W, W$\rightarrow$SA, SA$\rightarrow$W, and finally W$\rightarrow$S. 

\begin{figure}[t]
    \centering
    \includegraphics[width=1\linewidth]{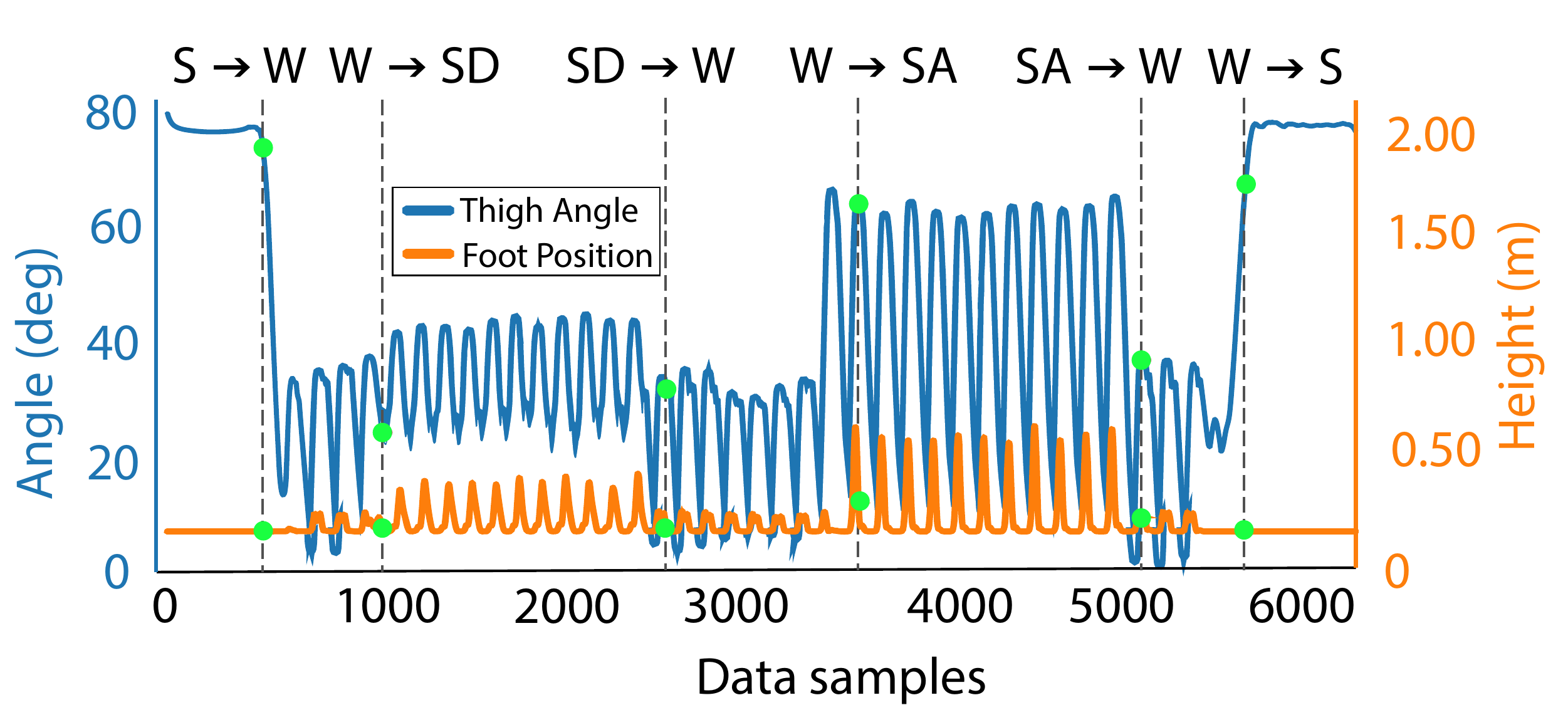}
    \vspace{-0.4cm}
\caption{Collected data with IMU-based motion capture system and detected transitions}
\label{fig:XsensResults}
\end{figure}

The collected data was input into the FSM in an offline way to analyze the performance of ICF detection and classification for each transition. The Fig.~\ref{fig:XsensResults} shows that for sit-to-walk, walk-to-stair descend, stair-descend-to-walk, walk-to-stair ascend, stair ascend-to-walk, and walk-to-sit transitions ICFs are detected, and all of the transitions are correctly classified. 

This evaluation also indicated that to detect the transitions W$\rightarrow$S and S$\rightarrow$W the changing ICF-3 from when $\theta = 10$ to while $\theta$ is in between 70-75 would increase the detection accuracy performance.

\subsubsection{Onboard Bench-marking}

\begin{figure*}[t]
    \centering
    \includegraphics[width=1\linewidth]{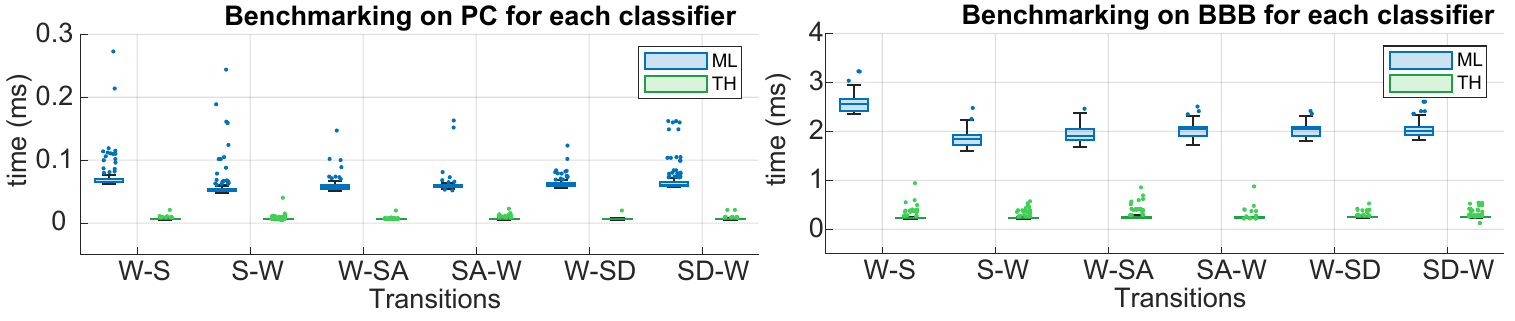}
    \vspace{-0.6cm}
\caption{Benchmarking results on two devices with the methods as machine learning (LR and SVM) and TH implementation for each classifier separately}
\label{fig:BenchmarkingwSingle}
\end{figure*}

The computational costs of the TH method and the ML-based approach are evaluated and compared on the BeagleBone Black. This embedded single-board computer runs on a Debian-based Linux distribution that is specifically optimized for embedded systems, ensuring efficient performance.

The collected data with the IMU-based motion capture system is sequentially inputted to the FSM. When the binary classification is performed, the time spent is recorded for each individual classifier and also for the FSM which includes the detection of ICF, and extraction of the corresponding feature as well. This process is performed for both of the methods with 100 transition cycles for each classifier. 

To show the importance of efficient onboard computing, the test is repeated on both a notebook computer (Apple M1 processor, 8-core CPU, 7-core integrated GPU with 8 GB RAM) and the embedded computer (BeagleBone Black, 1GHz ARM® Cortex-A8, 512MB RAM). The results of the benchmarking tests for the two methods are illustrated in Fig.~\ref{fig:BenchmarkingwSingle}.

\subsection{Experimental Protocol with Human Subjects}

\begin{figure}[b]
    \centering
    \includegraphics[width=0.7\linewidth]{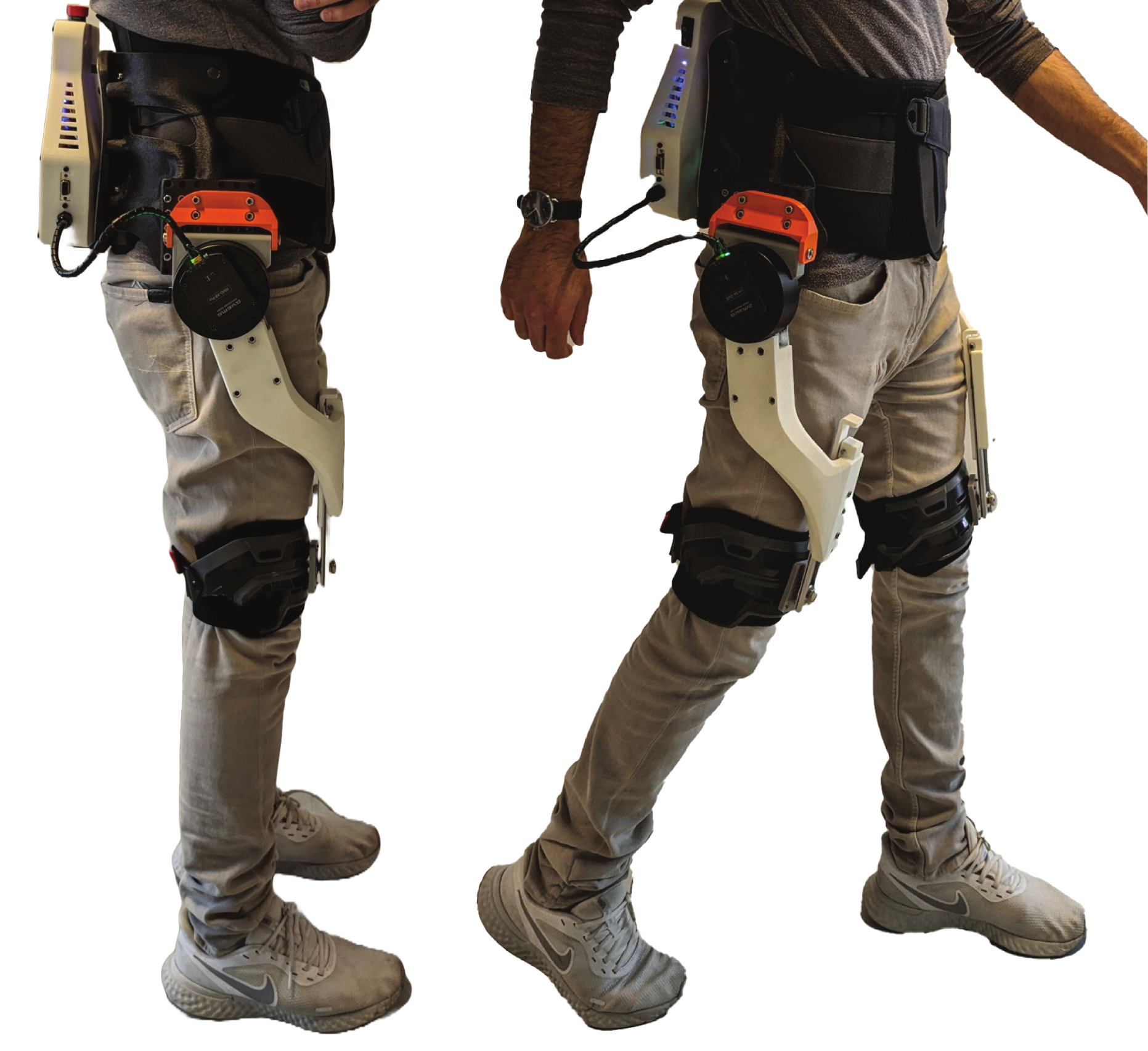}
    \vspace{-0.2cm}
\caption{The active hip orthosis eWalk.}
\label{fig:exophoto}
\end{figure}

The hip orthosis eWalk is designed for partial assistance for the hip flexion/extension movement in collaboration between the EPFL research group Biorob-REHAssist and the company Sonceboz. eWalk has 2 active degrees of freedom which are actuated with DC servo motors (Gyems RMD X8 Pro), as shown in Fig.\ref{fig:exophoto}. To allow natural walking, the abduction/adduction and the internal rotation of the hip are not constrained to allow natural walking due to the compliant nature of thigh segments.

The position is measured by an incremental encoder on the motor side for the assistive device. The zero position of the joints is calibrated by the homing technique. Ground reaction forces are measured by the flexible insole based on force-sensing resistors. eWalk has an embedded computer (BeagleBone Black, Texas Instruments, USA) that collects and keeps the log of all the sensory data. Wireless communication with both devices is established through a Wi-Fi module.

During the experiments, eWalk is utilized in passive mode due to the transparent nature of the device with low mass and low friction. 

During the tests, ten healthy subjects with the age of $25.6 \pm 2.76$ years, a height of $178.4 \pm 8.1$ cm, and a body mass of $71.9 \pm 9.68$ kg were asked to perform the trial. 

The experimental scenario included four locomotion modes (level-ground walking, stair ascent, stair descent, and sit), and six locomotion transitions were tested in the environment depicted in Fig~\ref{fig:OnlineTestsOutdoorEnvironment}. Locomotion transitions as W $\rightarrow$ SA, SA $\rightarrow$ W, W $\rightarrow$ S,  S $\rightarrow$ W, W $\rightarrow$ SD and SD $\rightarrow$ W are performed in consecutive order during the experiment. There were 15-step stairs. For all locomotion modes, subjects were instructed to walk at their self-selected walking speed. The sitting position on the bench is self-selected. Each trial is repeated 5 times by each subject. Therefore each locomotion transition is tested 50 times in total.

\subsubsection{Data Labeling}
The ground truth data is labeled during the experimental procedure. The person who is observing the subject during the trials puts the ground truth label for each transition and steady-state locomotion type. This label is logged in real-time together with the rest of the data of the exoskeletons.

\subsubsection{Performance of classifiers for transitions}
\begin{figure}[h]
    \centering
    \includegraphics[width=0.95\linewidth]{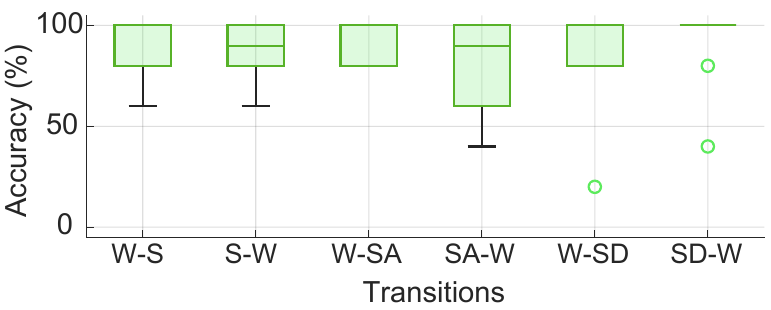}
    \vspace{-0.2cm}
\caption{The results real-time recognition accuracy (\%) of locomotion transitions from experiments with eWalk exoskeletons}
\label{fig:RecognitionAccuracyTransitions}
\end{figure}

The recognition accuracy results are reported in Fig.~\ref{fig:RecognitionAccuracyTransitions} for the experiments with the eWalk exoskeleton. The recognition accuracy is defined as shown in Eqn.~\ref{eq:transitionAccuracy}:

\vspace{-0.2cm}
\begin{equation}
\label{eq:transitionAccuracy}
A = \frac{ N_{\text{correctly detected transitions}}}{ N_{\text{total transitions}}}.100
\end{equation}
\vspace{-0.3cm}

where $N_\text{correctly detected transitions}$ is the detected transitions at the correct moment  and $N_\text{total transitions}$  was the number of total trials. If the method does not recognize the transition at the time of transition, it is considered not detected, even if the steady state is correctly classified. 

\smallskip
\section{Discussion}
\label{sec:Discussion}


\subsection{The importance of real-time performance}

Classification delays, while possibly acceptable for health monitoring applications, can become problematic in real-time control scenarios \cite{Cheng2021}. In these scenarios, where locomotion mode classification is performed onboard in real-time alongside a high-level controller, the computational load might not only affect transition detection latency but may also influence controller performance.

Haptic systems typically utilize a local 1 kHz control loop to overcome device dynamics and display high-frequency haptic feedback to the user \cite{Navarro2018}.  In physical human-robot interaction (pHRI), for the robot to react rapidly in the case of an impact or to be as transparent as possible when physically collaborating with a human, its control loop should run at a minimum of 1 kHz \cite{Steinbach2012}. 

Given these considerations and the inherent limitations of computational resources in embedded systems, it becomes imperative to seek methods that can minimize computational costs while upholding classification accuracy. Note that the accuracy outcomes of ML-based and TH methods are identical as long as the selected ML model's output values are utilized as the threshold values. Consequently, TH methods demonstrate a notable reduction in computational costs, as presented in Fig.~\ref{fig:BenchmarkingwSingle}. For example, for the W$\rightarrow$S classifier, the median value of computational time changes from 0.2273 ms to 2.568 ms which is about making the threshold TH method 11 times faster. 

The computational costs in terms of time for both of the methods on a notebook computer are quite low. However, when assessing computational performance on an embedded board, the TH method outperforms the ML-based approach in terms of computational time, showcasing its efficiency in resource-constrained settings.

\subsection{Verification of the system}

As presented in Table~\ref{tab:RealTimeAcuracy}, when all the data points are provided to FSM one by one the overall accuracy is between $80\% - 99.90\%$ in all of the cases. Since the dataset \cite{Camargo2021} does not include W$\rightarrow$S, and S$\rightarrow$W transitions, they couldn't be reported. 

Our experiments with the Xsens measurement system yielded highly promising results as presented in Section\ref{sec:Experiments}. It's noteworthy that all transitions were accurately classified, showcasing the robustness of our approach. However, a subtle point deserves attention, as depicted in Fig.~\ref{fig:XsensResults}: the transition from W $\rightarrow$ SA exhibited a one-step delay.

It's essential to clarify that this delay isn't indicative of a misclassification; rather, it stems from the methodology employed. In this particular test, we didn't measure ground reaction forces. Instead, we relied on foot position in the vertical axis to detect load and unload conditions. Consequently, the observed one-step delay is a reflection of the limitations in acquiring foot position data, as illustrated in Fig.~\ref{fig:XsensResults}.

These results collectively underscore the versatility and robustness of the proposed method, further validated by its successful implementation across three distinct measurement systems.

\subsection{Real-time results with assistive devices}

During the real-time locomotion recognition experiments, high accuracy is reported as illustrated in Fig.~\ref{fig:RecognitionAccuracyTransitions}. This result shows that the proposed strategy could be utilized in real-time recognition with high accuracy and low computational costs.

\begin{figure}
    \centering
    \includegraphics[width=0.90\linewidth]{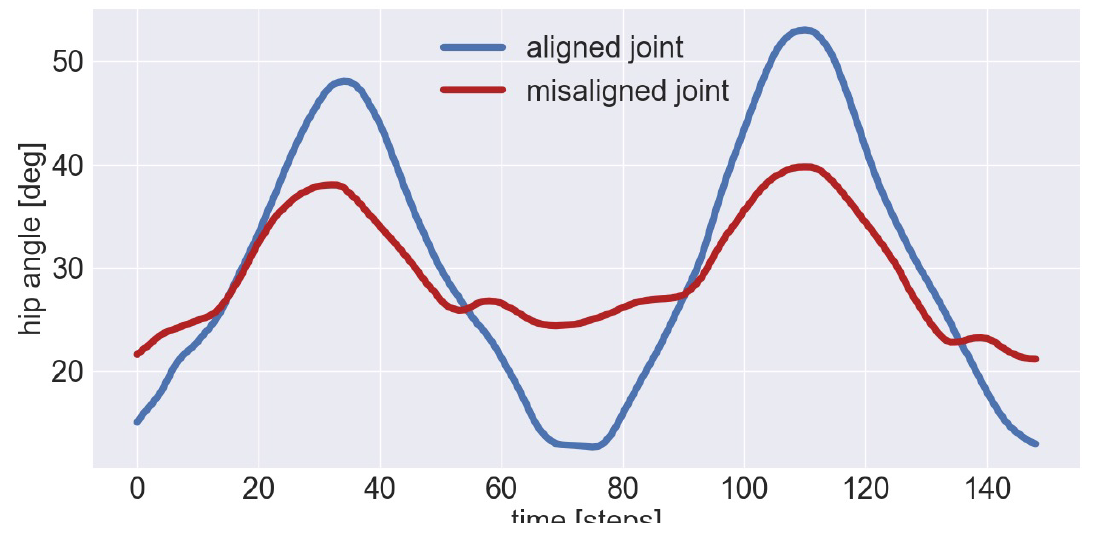}
    \vspace{-0.4cm}
\caption{Thigh angle trajectories of measured with during W$\rightarrow$SD when the exoskeleton joints are misaligned and aligned properly.}
\label{fig:eWalkcycles}
\end{figure}

Although high accuracy is reported, the recognition accuracy in other scenarios outperforms the results in SA$\rightarrow$W and  W$\rightarrow$SD during the experiments with eWalk. The main reason behind this is the proper alignment of the hip flexion/extension joint to the user's joint alignment. Fig. \ref{fig:eWalkcycles} provide an example for a W$\rightarrow$SD case. Since the joint positions of the exoskeleton were placed above the body ones due to the width of the trunk, the thigh angle measurement was smaller than the actual one. Therefore the range of motion of the joint is decreased with a smaller peak. Therefore, the feature ($ICF-1 = \theta_{th,MHF} - \theta_{th,HS}$) never overtakes the threshold for the transition detection. 

As stated in \cite{Cheng2022, Cheng2021}, activity mode misclassifications most frequently occur between walking and walk-to-stair transitions. To be able to reduce these misclassifications, a conditional offset is suggested for steady-state modes of walking and steady-state stairs in \cite{Cheng2022}.

\subsection{Limitations}
Firstly, the open-source dataset used for training the classifiers was collected from subjects who were not wearing a wearable assistive device. This limits our understanding of how well the model can adapt to different levels of assistance, as the kinematics of users can vary depending on the level of assistance received. Additionally, since the measurements were obtained from the sensors of the exoskeleton, any misalignment between human joints and exoskeleton joints can impact the reliability of the data. Moreover, while the proposed approach demonstrates satisfactory results with a subject-independent model, it is important to consider that the accuracy of the implementation may deteriorate when applied to target populations using exoskeletons, as their gait dynamics may differ from those of healthy users.

\section{Conclusion}
\label{sec:Conclusion}

In conclusion, our study highlights the significance of enhancing real-time embedded performance through a threshold-based approach, achieving an efficient balance between computational costs and high classification accuracy. The validation of utilized finite-state machine using data from diverse measurement systems shows its versatility and reliability.

Furthermore, our approach's practicality is demonstrated in the context of lower-limb wearable robots, as evidenced by human subject experiments involving an assistive device and a total of 10 healthy subjects. These results not only highlight the method's effectiveness but also its potential to significantly impact the field of assistive technology. Future work should focus on the verification of the suggested method with a target population of the exoskeletons and also should incorporate a control strategy that facilitates seamless human-robot interaction with the assistance of the exoskeleton.

\bibliographystyle{IEEEtran}
\bibliography{bibliography.bib}

\newpage

\listoftodos

\end{document}